\documentclass[11pt]{article}

\usepackage[final]{acl}

\usepackage{times}
\usepackage{latexsym}
\usepackage{multirow}
\usepackage{amsmath}
\usepackage{booktabs}
\usepackage[T1]{fontenc}

\usepackage[utf8]{inputenc}

\usepackage{microtype}

\usepackage{inconsolata}

\usepackage{graphicx}

\title{When AI Settles Down: Late-Stage Stability as a Signature of AI-Generated Text Detection}

\author{
  Ke Sun\thanks{This work was done when Ke Sun was a visiting student at Westlake University.}, 
  Guangsheng Bao, Han Cui, Yue Zhang\thanks{Corresponding author.} \\
  School of Engineering, Westlake University, China
}

\begin{document}
\maketitle
\begin{abstract}
Zero-shot detection methods for AI-generated text typically aggregate token-level statistics across entire sequences, overlooking the temporal dynamics inherent to autoregressive generation. We analyze over 120k text samples and reveal \textit{Late-Stage Volatility Decay}: AI-generated text exhibits rapidly stabilizing log probability fluctuations as generation progresses, while human writing maintains higher variability throughout. This divergence peaks in the second half of sequences, where AI-generated text shows 24--32\% lower volatility. Based on this finding, we propose two simple features: \emph{Derivative Dispersion} and \emph{Local Volatility}, which computed exclusively from late-stage statistics. Without perturbation sampling or additional model access, our method achieves state-of-the-art performance on EvoBench and MAGE benchmarks and demonstrates strong complementarity with existing global methods.
\end{abstract}

\section{Introduction}

The widespread adoption of large language models (LLMs) has raised significant concerns about the potential misuse of AI-generated text, making accurate detection of AI-generated content an increasingly important task \cite{yang2024survey,wu2025survey}. Recent zero-shot detection methods, such as DetectGPT \cite{mitchell2023detectgpt}, Fast-DetectGPT \cite{bao2024fastdetectgpt}, and LLR \cite{su2023detectllm} have achieved promising results by analyzing the probabilistic characteristics for different AI-generated and human written texts. These approaches typically compute global statistics, such as average model log-probability, rank, curvature, or sampling discrepancy over the entire text sequence.
\begin{figure}[t]
    \begin{center}
       \includegraphics[width=1\linewidth]{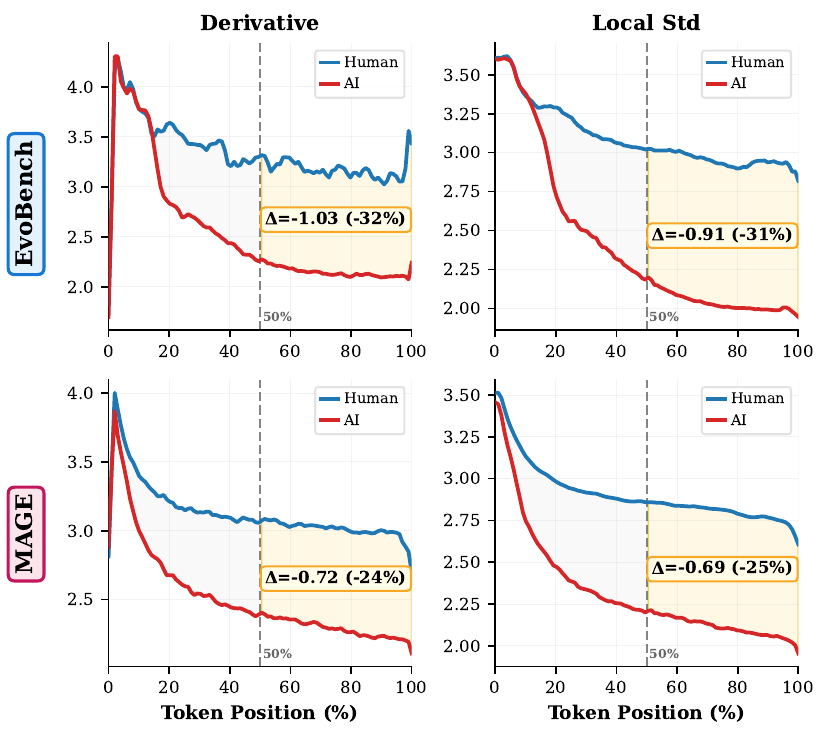}
    \end{center}
       \caption{ Temporal dynamics of \textbf{log probability} features for human vs. AI text on EvoBench (top) and MAGE (bottom). Left: Absolute derivative (change rate); Right: Local standard deviation (volatility). Shaded regions show the human-AI gap, with $\Delta$ indicating mean difference (AI $-$ Human) and relative percentage. Yellow regions (50--100\%) highlight larger divergence in the second half.
       }
    \label{fig:intro}
\end{figure}

Despite their effectiveness, these methods share a common implicit assumption: \emph{token-level statistics are treated as stationary and exchangeable across the sequence.} That is, discriminative signals are assumed to be uniformly distributed, so sequence-wide aggregation suffices. This assumption does not fully leverage characters of the autoregressive nature of LLM generation. Text is not produced all at once; instead, it is generated token by token, with each prediction conditioned on an ever-growing context. As a result, the statistical properties of generated tokens may evolve dynamically over time.


Recent empirical studies have observed that detection performance improves with increasing text length, with longer texts generally being easier to detect \cite{li2024mage,xu2024training,bao2024fastdetectgpt}. However, this phenomenon is typically attributed to longer texts simply ``containing more information.'' What remains largely unexplored is \emph{how discriminative signals are distributed across different stages of the generation process}, and whether later tokens exhibit qualitatively different behavior from earlier ones.


To address this issues, we conduct a large-scale temporal analysis of over 120K human-written and AI-generated texts from the EvoBench \cite{yu2025evobench} and MAGE \cite{li2024mage} benchmarks. Using multiple surrogate models, we examine how various token-level statistics including negative log-probability, rank, entropy, and sampling discrepancy -- evolve across normalized text positions. Crucially, we go beyond raw values and analyze two second-order temporal dynamics: the \emph{derivative}, which captures how rapidly a statistic changes between consecutive tokens, and the \emph{local standard deviation}, which measures regional volatility within a sliding window.


Our analysis reveals that the temporal dynamics of negative log-probability provide particularly discriminative signals. As illustrated in Figure~\ref{fig:intro}, we observe that both the derivative and local standard deviation of negative log-probability decrease more rapidly for AI-generated text than for human text, with the gap widening noticeably in the second half of the sequence (50\%--100\% positions). Quantitatively, on EvoBench, the derivative of negative log-probability for AI text is 32\% lower than that of human text ($\Delta=-1.03$), and local standard deviation is reduced by 31\% ($\Delta=-0.91$); on MAGE, similar reductions of 24\% and 25\% are observed. In other words, AI-generated tokens in the latter portion of text exhibit smaller and more stable probability fluctuations as the model's predictions become increasingly consistent as context accumulates. We term this pattern \textbf{Late-Stage Volatility Decay}. This phenomenon reflects the inherent behavior of autoregressive generation: as more context becomes available, the model's prediction distribution sharpens, leading to reduced variability in token-level statistics. Human writing, by contrast, continues to introduce unexpected lexical choices and maintains higher volatility throughout.

Based on this observation, we propose two simple yet effective temporal features computed from token negative log-probability to capture this late-stage dynamics: (1) \textit{derivative dispersion}, the standard deviation of the absolute first derivative in the second half, quantifying the intensity of probability fluctuations; and (2) \textit{local volatility}, calculated as the mean of local standard deviation within a sliding window over the second half, this metric measures local stability, capturing how AI text tends to maintain smoother, more consistent probability transitions as generation progresses.



Since our temporal cues are largely orthogonal to prior signals, a simple fusion with Fast-DetectGPT~\cite{bao2024fastdetectgpt} yields consistent gains by combining \textit{global distributional} and \textit{local temporal} evidence. Experiments on MAGE~\cite{li2024mage} and EvoBench~\cite{yu2025evobench} show that our approach excels on longer texts while remaining competitive on shorter ones, and that fusion further improves robustness across diverse generators. Our contributions are summarized as follows:


\begin{itemize}
    \item We systematically reveal the \textit{Late-Stage Volatility Decay} phenomenon in AI-generated text—a distinctive temporal signature arising from the autoregressive generation process.

    \item We propose a simple yet effective method that operationalizes this phenomenon through second-half temporal features, requiring no additional training or model access beyond a surrogate LLM.
    
    \item  We achieve state-of-the-art detection performance across multiple benchmarks, demonstrating the value of temporal analysis in machine-generated text detection.
\end{itemize}
\section{Temporal Dynamics of Zero-Shot Detection Metrics}

\label{sec:analysis}



To understand whether discriminative signals vary across text positions, we empirically analyze how these statistics evolve over generation to identify which temporal patterns best distinguish human and AI-generated text.

\subsection{Analysis Setup}
\label{subsec:setup}

\paragraph{Datasets and Models.}
We analyze over 120k text samples from two established benchmarks: EvoBench~\cite{yu2025evobench} and MAGE~\cite{li2024mage}. Both datasets contain parallel human-written and AI-generated texts across diverse domains and generation settings. We use Llama-3-8B-Instruct~\cite{grattafiori2024llama} as the primary surrogate model for computing token-level statistics. Results with alternative surrogate models GPT-J-6B~\cite{gpt-j} and Falcon-7B~\cite{almazrouei2023falcon} are provided in Appendix~\ref{app:surrogate}.
Additional experiments on frontier models including reasoning-enhanced systems are presented in Appendix~\ref{app:frontier}.

\begin{figure*}[t]
    \centering
    \includegraphics[width=1.0\linewidth]{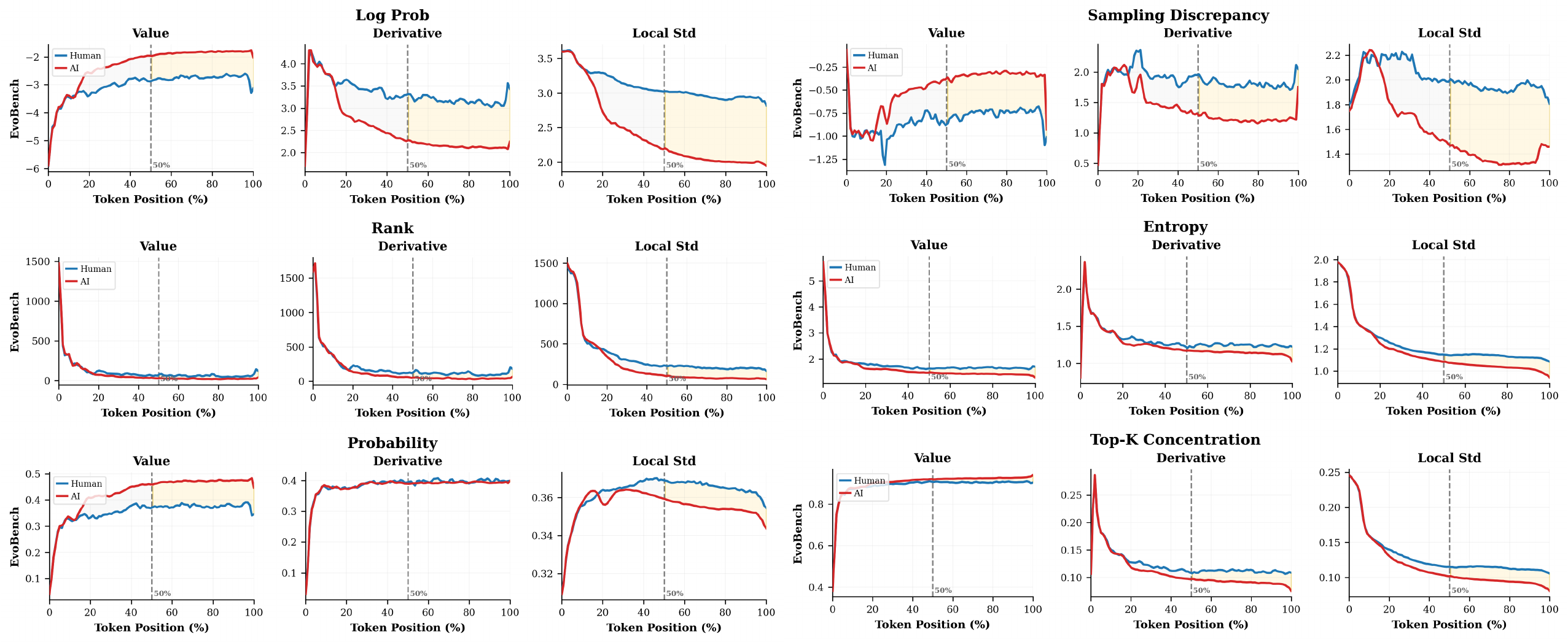}
    \caption{Temporal dynamics of six base metrics on EvoBench. Each row shows a metric's raw value (left), absolute derivative (middle), and local standard deviation (right) across normalized token positions (0--100\%). Log probability and Sampling Discrepancy (alias Conditional Probability Curvature) display clear human-AI separation that widens in the second half, while entropy and top-$k$ concentration show minimal divergence.}
    \label{fig:analysis}
\end{figure*}

\paragraph{Zero-Shot Detection Metrics.}
We examine seven token-level statistics commonly used in zero-shot detection: (1)~\textit{log probability} $\log P(x_t|x_{<t})$, reflecting model confidence at each position; 
(2)~\textit{Sampling Discrepancy}~\cite{bao2024fastdetectgpt}, referred to as conditional probability curvature in the original work;
(3)~\textit{rank} of the actual token among all vocabulary items;
(4)~\textit{entropy} of the predicted distribution;  
(5)~\textit{token probability} $P(x_t|x_{<t})$ the raw conditional probability without log transformation; and (6)~\textit{top-$k$ concentration}, the cumulative probability of the top-10 tokens.

\paragraph{Temporal Features.}
To capture the dynamics of these metrics, we compute two second-order features at each position: the \textit{absolute first derivative} $|m_t - m_{t-1}|$, measuring the rate of change between consecutive tokens, and the \textit{local standard deviation} within a sliding window of 20 tokens, quantifying regional volatility. For cross-sample comparison, we normalize token positions to percentiles (0--100\%), enabling aggregation across texts of varying lengths.

\subsection{Comparative Metric Sensitivity}
\label{subsec:metric_sensitivity}

We first investigate which base metrics exhibit meaningful temporal differences between human and AI-generated text. Figure~\ref{fig:analysis} presents the temporal evolution of all six metrics on EvoBench, showing the raw values (left), absolute derivatives (middle), and local standard deviations (right) across normalized text positions.
\paragraph{Not All Metrics Capture Temporal Dynamics.}
We first observe that temporal discriminability varies dramatically across metrics. While \textit{log probability} and \textit{sampling discrepancy} show clear separation between human and AI curves, particularly in their derivative and local standard deviation, metrics such as \textit{entropy}, \textit{token probability}, and \textit{top-$k$ concentration} exhibit nearly overlapping trajectories throughout the sequence.


To quantify these differences, we measure the slope ratio: how much steeper AI's temporal trend is compared to human. As shown in Table~\ref{tab:metric_temporal}, log probability exhibits the strongest temporal dynamics, with slope ratios of $2.4$ for derivative and $2.6$ for local standard deviation on EvoBench, and $2.4$ for derivative and $1.8$ for local standard deviation on MAGE. Sampling discrepancy shows comparable trends, with ratios of $2.2$ (derivative) and $2.5$ (local) on EvoBench, and $2.3$ (derivative) and $2.2$ (local) on MAGE, though it requires additional sampling overhead. In contrast, entropy, token probability, and top-$k$ concentration show slope ratios close to $1.0$, indicating minimal temporal differentiation. Based on these findings, we focus subsequent analysis on log probability, which provides the strongest temporal signal without additional computation.


\begin{table}[t]
\centering
\small
\setlength{\tabcolsep}{3pt}
\begin{tabular}{lcccc}
\toprule
\multirow{2}{*}{\textbf{Metric}} & \multicolumn{2}{c}{\textbf{EvoBench}} & \multicolumn{2}{c}{\textbf{MAGE}} \\
\cmidrule(lr){2-3} \cmidrule(lr){4-5}
 & Deriv. & Local & Deriv. & Local \\
\midrule
Log Probability & ${2.4}$ & ${2.6}$ & ${2.4}$ & ${1.8}$ \\
Sampling Discrepancy & $2.2$ & $2.5$ & $2.3$ & $2.2$ \\
Rank & $1.2$ & $1.2$ & $1.0$ & $1.1$ \\
Entropy & $1.4$ & $1.3$ & $1.4$ & $1.2$ \\
Token Probability & $0.9$ & -- & -- & -- \\
Top-$k$ Concentration & $1.4$ & $1.3$ & $0.9$ & $1.0$ \\
\bottomrule
\end{tabular}
\caption{Slope ratio across metrics, measuring how many times steeper AI's temporal trend is compared to human. Higher values indicate stronger temporal dynamics. ``--'' indicates inconsistent trend directions between human and AI.}
\label{tab:metric_temporal}
\end{table}

\subsection{Late-Stage Volatility Decay}
\label{subsec:late_stage}



We begin by examining the temporal dynamics of log probability, which emerges as the most informative metric throughout our analysis. Figure~\ref{fig:intro} visualizes the derivative and local standard deviation of log probability across normalized token positions for human-written and AI-generated text.

Two consistent trends are immediately apparent. First, for both human and AI-generated, volatility decreases as generation progresses, reflecting increasing contextual constraint. Second, and more importantly, AI-generated text stabilizes substantially faster than human text. The decline in both derivative magnitude and local variability is markedly steeper for AI-generated sequences, and the gap between the two curves widens progressively over time.

This divergence becomes most pronounced in the second half of the sequence (50\%–100\% positions). In this region, AI-generated text exhibits significantly smaller fluctuations in token-level log probability, while human text maintains higher volatility. We refer to this systematic asymmetry as \textit{Late-Stage Volatility Decay}.

\paragraph{Quantitative Characters.}
Table~\ref{tab:late_stage} quantifies the late-stage volatility decay from two complementary perspectives. The \textit{decay rate} measures how fast each class stabilizes from the first to second half. The \textit{second-half gap} measures the relative difference between human and AI-generated text in later positions, directly reflecting discriminability.

On EvoBench, human text decays by only 9.6\% from the first to second half in terms of derivative, while AI-generated text decays by 27.1\%, a rate $2.4$ times faster. This asymmetric decay widens the human-AI distinction: in the second half, the relative difference reaches 32\%. The local standard deviation exhibits the same pattern, where $2.6$ times faster AI decay leads to a 31\% gap in the second half.


\begin{table}[t]
\centering
\small
\setlength{\tabcolsep}{3.0pt}
\begin{tabular}{llcccc}
\toprule
\textbf{Dataset} & \textbf{Feature} & \textbf{H-decay} & \textbf{A-decay} & \textbf{Ratio} & \textbf{Gap\%} \\
\midrule
\multirow{2}{*}{EvoBench} 
  & Derivative & $-9.6$\% & $-27.1$\% & $2.4$ & $-32$\% \\
  & Local Std & $-8.9$\% & $-26.5$\% & $2.6$ & $-31$\% \\
\midrule
\multirow{2}{*}{MAGE} 
  & Derivative & $-7.5$\% & $-16.7$\% & $1.9$ & $-24$\% \\
  & Local Std & $-7.3$\% & $-16.5$\% & $1.9$ & $-25$\% \\
\bottomrule
\end{tabular}
\caption{Late-stage volatility decay of log probability features. \textit{H-decay}/\textit{A-decay}: percentage decrease from first to second half for human/AI-generated text. \textit{Ratio}: how many times faster AI decays than human. \textit{Gap\%}: relative human-AI difference in the second half. 
AI-generated text consistently decays $1.9$--$2.6$ faster across datasets.}
\label{tab:late_stage}
\end{table}

These patterns generalize across datasets. On MAGE, AI-generated text decays approximately $1.9$ times faster than human text, resulting in 24--25\% second-half gaps for both features. While absolute magnitudes vary between datasets, the qualitative pattern remains consistent.

\paragraph{Interpretation.}
This phenomenon reflects the autoregressive nature of LLM generation. As more context accumulates, the model's prediction distribution becomes increasingly constrained, leading to more confident outputs. Consequently, token surprise $s_t = -\log P(x_t|x_{<t})$, which measures how unexpected each token is, exhibits smaller fluctuations in later positions~\cite{basani2025diversity,hou2025theoretical}. Both the rate of change and regional variability of surprise diminish as the model's certainty grows.
Human writing does not follow this optimization process. Writers introduce unexpected word choices and stylistic variations throughout their text, maintaining higher surprise volatility even in later portions. This fundamental difference manifests as the late-stage volatility decay pattern and motivates our focus on second-half statistics for detection.


\subsection{Summary of Findings}




Our empirical analysis demonstrates that AI-generated text exhibits accelerated stabilization in later stages of generation, leading to significantly reduced volatility in token-level surprise. This reveals that discriminative signals are not uniformly distributed across positions, but instead concentrate in the second half of sequences due to this late-stage stabilization. As a result, features computed from the second half capture substantially stronger human–AI divergence than full-sequence averages, which dilute informative late-stage signals with less discriminative early tokens. This insight directly motivates the design of our detection method.

\section{Method}
\label{sec:method}



Motivated by our empirical analysis in Section \ref{sec:analysis}, we build a simple, training-free method that explicitly exploits this late-stage temporal asymmetry. Rather than aggregating statistics over the entire sequence, we focus on the second half of the text, where human and AI divergence is maximal.

\subsection{Late-Stage Stability Metrics}
Following \citet{bao2024fastdetectgpt}, we adopt a surrogate LLM for calculating token-level probabilities. Given a text $x=(t_1,\dots,t_n)$, we query the surrogate model to obtain $p_i = P(t_i \mid t_{<i})$ and define token surprise $s_i=-\log p_i$.
Based on the Late-Stage Volatility Decay phenomenon, we focus exclusively on the \emph{second half} of the sequence to characterize late-stage stability.
Let $\mathcal{H}_2=\{i \mid i > \lfloor n/2 \rfloor\}$ denote the set of token positions in the second half. Motivated by the two temporal features examined in Section~\ref{sec:analysis}, we define corresponding detection features that directly quantify their dispersion in the second half.

\paragraph{Derivative Dispersion (DD).}

This feature captures the variability of surprise changes between consecutive tokens. We first compute the absolute first-order difference of surprise,
\begin{equation}
d_i = \left|s_i - s_{i-1}\right|,\quad i=2,\dots,n,
\end{equation}
which measures how rapidly surprise fluctuates from one token to the next. We then quantify the dispersion of these change rates in the second half:
\begin{equation}
\mathrm{DD} = \mathrm{Std}\big(\{d_i\}_{i\in \mathcal{H}_2}\big).
\end{equation}
AI-generated text, with its increasingly stable predictions, exhibits lower DD values as the change rates become more uniform in later positions.


\paragraph{Local Volatility (LV).}
To capture regional stability patterns that may not be reflected in global statistics, we compute a sliding-window standard deviation around each position:
\begin{equation}
\ell_i = \mathrm{Std}\big(s_{i-w/2},\dots,s_{i+w/2}\big),
\end{equation}
with window size $w$ (default $w{=}20$). This measures how much surprise varies within a local neighborhood. We then average these local volatility values over the second half:
\begin{equation}
\mathrm{LV} = \mathrm{Mean}\big(\{\ell_i\}_{i\in \mathcal{H}_2}\big).
\end{equation}
This feature is particularly sensitive to the smooth, consistent probability transitions characteristic of AI-generated text in later positions.

\subsection{Combined Detection Metric}
Both DD and LV features measure different aspects of volatility and yield lower values for AI-generated text due to late-stage stabilization. We negate and sum them to form our Temporal Stability Detection (TSD) score:
\begin{equation}
S_{\text{TSD}} = -(\mathrm{DD} + \mathrm{LV}).
\end{equation}
Higher $S_{\text{TSD}}$ indicates greater likelihood of AI-generated text. Classification is performed by thresholding: $\hat{y}={I}[S_{\text{TSD}}>\tau]$, where $\tau$ is determined on validation data.

\subsection{Fusion with Global Metrics}
Our temporal features capture \textit{when} statistical patterns emerge (late-stage dynamics), which is orthogonal to existing methods that focus on \textit{what} distributional anomalies exist (sequence-wide aggregation). This complementarity suggests a natural fusion strategy: combining temporal features with global statistical methods to leverage both perspectives.

We demonstrate this with Fast-DetectGPT~\cite{bao2024fastdetectgpt}, a representative global method that computes sampling discrepancy across the entire sequence. A simple additive fusion combines both signals:
\begin{equation}
S_{\text{fusion}} = S_{\text{TSD}} + S_{\text{global}}.
\end{equation}

By explicitly modeling late-stage volatility, our method departs from the common assumption of stationarity in token-level statistics. 
The resulting features are simple, interpretable, and computationally efficient, yet capture a fundamental temporal signature of machine-generated text.
\section{Experiments}
\label{sec:experiments}
\begin{table*}[t]
\centering
\small
\setlength{\tabcolsep}{3.5pt}
\begin{tabular}{ll|ccccccc|c|c}
\toprule
\multirow{2}{*}{\textbf{Category}} & \multirow{2}{*}{\textbf{Method}} & \multicolumn{8}{c|}{\textbf{EvoBench}} & \textbf{MAGE} \\
\cmidrule(lr){3-10} \cmidrule(lr){11-11}
 & & LLaMA-3 & LLaMA-2 & Claude & GPT-4o & Gemini & GPT-4 & Qwen & Avg & Avg \\
\midrule
\multirow{6}{*}{Global} 
& Likelihood & 88.84 & 61.68 & 79.70 & 77.80 & 78.89 & 77.80 & 77.40 & 77.44 & 56.81 \\
& Entropy & 28.85 & 57.58 & 33.61 & 35.43 & 38.34 & 38.10 & 30.34 & 37.46 & 51.90 \\
& Rank & 68.49 & 58.38 & 64.50 & 64.64 & 64.46 & 63.43 & 55.06 & 62.71 & 58.99 \\
& Log-Rank & 86.47 & 59.99 & 77.37 & 76.80 & 75.99 & 74.87 & \textbf{76.10} & 75.37 & 55.91 \\
& LLR & 66.46 & 49.61 & 61.73 & 60.57 & 59.12 & 58.43 & 63.08 & 59.86 & 51.78 \\
& Fast-Detect & \underline{94.17} & 80.27 & 81.57 & 74.89 & 80.48 & 75.58 & 71.96 & 79.85 & 69.69 \\
\midrule
\multirow{4}{*}{Temporal} 
& Lastde & 90.80 & 80.54 & 76.75 & 71.34 & 76.50 & 71.51 & 69.20 & 76.66 & 70.71 \\
& FourierGPT & 63.99 & 57.23 & 58.20 & 55.52 & 56.69 & 55.16 & 64.39 & 58.74 & 60.34 \\
& Diveye & 84.65 & 73.46 & 74.21 & 73.32 & 74.06 & 72.28 & 68.85 & 74.40 & 69.60 \\
& UCE & 75.62 & 61.35 & 68.72 & 62.04 & 61.28 & 59.38 & 63.63 & 64.57 & 54.17 \\
\midrule
\multirow{4}{*}{Ours} 
& DD & 91.12 & 80.77 & 82.56 & 80.57 & 79.86 & 78.48 & 71.78 & 80.73 & \underline{71.72} \\
& LV & 92.57 & 79.79 & 82.51 & \underline{84.88} & \underline{84.64} & \underline{83.58} & \underline{74.55} & 83.22 & 71.26 \\
& TSD & 92.33 & \underline{80.91} & \underline{85.20} & 84.55 & 84.06 & 82.93 & 73.58 & \underline{83.36} & 71.56 \\
& TSD+ & \textbf{94.50} & \textbf{83.42} & \textbf{87.12} & \textbf{85.68} & \textbf{87.40} & \textbf{85.00} & 74.48 & \textbf{85.37} & \textbf{75.20} \\
\bottomrule
\end{tabular}
\caption{Detection performance (AUROC \%) on EvoBench and MAGE. DD: Derivative Dispersion; LV: Local Volatility; TSD: DD + LV; TSD+: TSD fused with Fast-DetectGPT. Best results are \textbf{bolded}, second best are \underline{underlined}. Full MAGE results in Appendix~\ref{app:mage_full}.}
\label{tab:main_results}
\end{table*}

\subsection{Experimental Setup}

\paragraph{Evaluation Benchmarks.}
We conduct experiments on two complementary benchmarks. \textbf{EvoBench}~\cite{yu2025evobench} specifically targets the evolving nature of LLMs, comprising over 32,000 pairs across seven model families (GPT-4, GPT-4o, Claude, Gemini, LLaMA-2, LLaMA-3, Qwen) with 29 distinct model versions. EvoBench captures both version updates within model families and variations from fine-tuning or quantization, providing a rigorous test for detection robustness as source models evolve.
\textbf{MAGE}~\cite{li2024mage} provides broad coverage with over 30,000 test pairs spanning eight model families (including OpenAI GPT series, LLaMA, OPT, FLAN-T5, and others), representing diverse generation settings across multiple domains.

\paragraph{Baselines.}
We compare against representative zero-shot detection methods spanning multiple paradigms: (1)~\textit{Token-level statistics}: Likelihood~\cite{solaiman2019release}, Entropy~\cite{gehrmann2019gltr}, Rank~\cite{gehrmann2019gltr}, and Log-Rank~\cite{mitchell2023detectgpt} directly aggregate token-level statistics over the entire sequence; (2)~\textit{Relative scoring}: LLR (Log-Likelihood Ratio)~\cite{su2023detectllm} normalizes scores through cross-model comparison; (3)~\textit{Perturbation-based}: Fast-Detect~\cite{bao2024fastdetectgpt} efficiently estimates local curvature through conditional sampling; (4)~\textit{Sequential analysis}: Lastde~\cite{xu2024training} mines discriminative local subsequences from probability time series, FourierGPT~\cite{xu2024detecting} leverages frequency-domain analysis of token statistics, and Diveye~\cite{basani2025diversity} quantifies diversity patterns in surprisal fluctuations; (5)~\textit{Uncertainty Contraction Estimator}: UCE~\cite{hou2025theoretical} exploits uncertainty characteristics of model predictions. All baselines use the same surrogate model as our method for fair comparison.

\paragraph{Implementation Details.}
We use Llama-3-8B-Instruct as the surrogate model for all methods. Input sequences are truncated to a maximum of 512 tokens. For our temporal features, we use the second half of each sequence (starting at position $\lfloor n/2 \rfloor$) and a sliding window size of $w{=}20$ tokens for local volatility computation. TSD+ represents our method fuse with Fast-DetectGPT, which consistent with the original implementation~\cite{bao2024fastdetectgpt}. Following standard practice~\cite{mitchell2023detectgpt,bao2024fastdetectgpt}, we report AUROC as the primary metric,

\subsection{Main Results}

Table~\ref{tab:main_results} presents detection performance across both benchmarks. We report results for individual generator families on EvoBench and average performance on MAGE, with per-family results in Appendix~\ref{app:mage_full}.

Our simple temporal features achieve state-of-the-art performance among standalone methods, with TSD attaining 83.36\% on EvoBench and 71.56\% on MAGE, outperforming all baselines including Fast-DetectGPT. This is notable given our method's simplicity: we compute only second-order statistics from the second half of sequences, without perturbation sampling or frequency-domain transformations.

Compared to global methods, our approach shows pronounced advantages on recent, capable models. Specifically, On GPT-4o and GPT-4, TSD outperforms Fast-DetectGPT by 9.66\% and 7.35\%, respectively. Advanced LLMs produce fluent text with fewer global distributional anomalies, yet still exhibit late-stage volatility decay due to autoregressive generation. Compared to other sequential methods, our focused design also proves more effective: Lastde achieves 76.66\% through subsequence mining, Diveye reaches 74.40\% via diversity analysis. These methods capture general sequential structure but do not specifically target late-stage stabilization.


Fusion with Fast-DetectGPT yields further gains: TSD+ achieves 85.37\% on EvoBench and 75.20\% on MAGE. The improvement is especially pronounced on MAGE, whose generators are both architecturally diverse and highly variable in output length. In particular, FLAN-T5 samples are typically shorter, leaving fewer late-stage tokens and thus weakening the temporal cues that our stability-based features rely on (Detailed in Appendix~\ref{app:mage_length}). Fast-DetectGPT, by contrast, provides a more length-robust global discrepancy signal, which compensates for this short-text regime and stabilizes performance across encoder--decoder and decoder-only families (Appendix~\ref{app:mage_full}). Overall, the fusion alleviates length-induced failure modes and demonstrates the complementary nature of temporal dynamics and global detection signals.

\subsection{Ablation Study}

\begin{table}[t]
\centering
\small
\setlength{\tabcolsep}{2.5pt}
\begin{tabular}{l|cc|cc|cc}
\toprule
\multirow{2}{*}{\textbf{Position}} & \multicolumn{2}{c|}{\textbf{DD}} & \multicolumn{2}{c|}{\textbf{LV}} & \multicolumn{2}{c}{\textbf{TSD}} \\
\cmidrule(lr){2-3} \cmidrule(lr){4-5} \cmidrule(lr){6-7}
 & Evo & MAGE & Evo & MAGE & Evo & MAGE \\
\midrule
First Half & 59.91 & 66.54 & 62.37 & 67.26 & 63.59 & 68.02 \\
Full Seq & 72.64 & 70.31 & 75.82 & 70.39 & 77.00 & 70.49 \\
Second Half & \textbf{80.73} & \textbf{71.71} & \textbf{83.21} & \textbf{71.25} & \textbf{83.36} & \textbf{71.56} \\
\bottomrule
\end{tabular}
\caption{Ablation on position selection.}
\label{tab:ablation_position}
\end{table}

\paragraph{Position Selection.}
To validate the importance of late-stage features, we compare three position selection strategies: using only the first half, the full sequence, and only the second half of text.
As shown in Table~\ref{tab:ablation_position}, second-half features substantially outperform alternatives across all metrics and datasets. On EvoBench, TSD computed from the second half achieves 83.36\%, compared to 77.00\% for the full sequence and only 63.59\% for the first half. The performance gap between second-half and first-half features exceeds 20 points on EvoBench, directly confirming that late-stage volatility decay provides the primary discriminative signal. Using the full sequence dilutes this signal with less informative early-stage statistics, resulting in intermediate performance. These results empirically validate our analysis in Section~\ref{sec:analysis}: the human-AI divergence in volatility features accumulates over the sequence and reaches its maximum in the latter portion, making second-half statistics the most effective choice for detection. Further experiments in Appendix~\ref{app:position} confirm that the relative midpoint (50\%) consistently outperforms both alternative relative positions and fixed absolute positions across datasets.

\paragraph{Base Metric Selection.}
We evaluate the effectiveness of different base metrics for computing temporal features, applying the same TSD computation to each metric.

\begin{table}[t]
\centering
\small
\setlength{\tabcolsep}{6pt}
\begin{tabular}{l|cc}
\toprule
\textbf{Base Metric} & \textbf{EvoBench} & \textbf{MAGE} \\
\midrule
Entropy & 53.52 & 55.24 \\
Probability & 61.69 & 55.37 \\
Top-10 Concentration & 53.06 & 55.66 \\
Rank & 76.06 & 62.98 \\
Sampling Discrepancy & 77.66 & 68.97 \\
\midrule
Log Probability (Ours) & \textbf{83.36} & \textbf{71.56} \\
\bottomrule
\end{tabular}
\caption{TSD performance with different base metrics.}
\label{tab:ablation_metric}
\end{table}

As shown in Table~\ref{tab:ablation_metric}, log probability substantially outperforms all alternatives, achieving 83.36\% on EvoBench compared to 77.66\% for sampling discrepancy and 76.06\% for rank. Entropy, probability, and top-10 concentration yield near-random performance below 62\%. This result aligns with our analysis in Section~\ref{sec:analysis}: log probability directly measures token surprise, which exhibits the clearest late-stage volatility decay pattern due to the autoregressive generation mechanism. Other metrics either lack sensitivity to temporal dynamics or capture less discriminative aspects of the prediction distribution.

\begin{figure}[t]
    \begin{center}
       \includegraphics[width=1\linewidth]{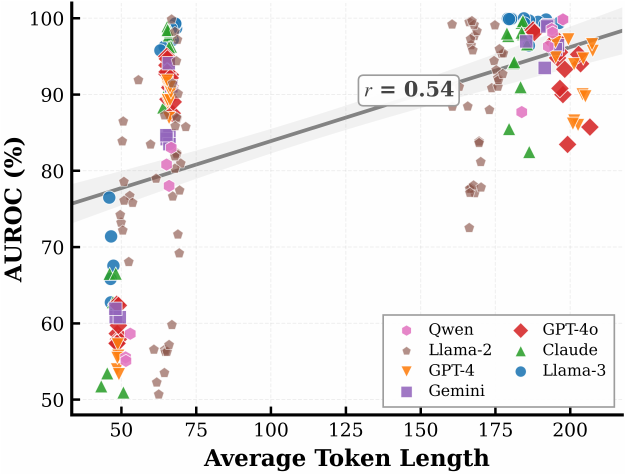}
    \end{center}
       \caption{ TSD performance (AUROC) versus average token length across EvoBench sources. Each point represents a dataset from one of seven model families. The dashed line shows linear regression fit with Pearson correlation $r = 0.54$.
       }
    \label{fig:length}
\end{figure}

\paragraph{Impact of Text Length.}
To investigate how text length affects our temporal features, we analyze performance across all seven model families in EvoBench. Figure~\ref{fig:length} plots AUROC against average token length for each source, revealing a clear positive correlation with Pearson coefficient $r = 0.54$.

The results demonstrate that our method excels on longer texts while showing limitations on shorter ones. For texts exceeding 150 tokens, TSD consistently achieves above 82\%, with many sources reaching 90\% or higher. In contrast, nearly all cases with performance below 70\% correspond to short texts with fewer than 100 tokens. This length dependency reflects the fundamental nature of Late-Stage Volatility Decay: longer sequences provide more tokens in the second half, allowing the human-AI divergence in volatility to fully manifest.

Short texts simply lack sufficient late-stage context for the stabilization pattern to emerge reliably. This trend is consistent with our findings on MAGE as well: across heterogeneous generator families, shorter outputs (e.g., FLAN-T5) exhibit systematically weaker temporal separability, and AUROC increases with average length, which detailed in Appendix~\ref{app:mage_length}. Together, EvoBench and MAGE provide aligned evidence that text length is a key factor governing the reliability of late-stage temporal cues.


\section{Related Work}

We review zero-shot detection for LLM-generated text, broadly grouping prior work into two streams: \textit{global statistical methods} that aggregate distributional cues over the whole text, and \textit{sequential/temporal methods} that exploit how token-level probability signals evolve along the generation trajectory.

\textbf{Global Statistical Detection Methods.}
Zero-shot detection leverages surrogate LLMs to extract distributional artifacts without requiring task-specific training data. Early approaches focused on absolute token statistics, employing metrics such as entropy, perplexity, and rank histograms to identify machine-generated anomalies~\cite{gehrmann2019gltr,solaiman2019release,ippolito2020bert}. To improve robustness, perturbation-based methods emerged, measuring the local curvature of the log-likelihood landscape around the candidate text, as exemplified by DetectGPT~\cite{mitchell2023detectgpt} and its more efficient variant Fast-DetectGPT~\cite{bao2024fastdetectgpt} and Glimpse~\cite{bao2025glimpse}. Concurrently, relative probability approaches have been developed to normalize scores through cross-model comparisons or likelihood ratios, including DetectLLM~\cite{su2023detectllm}, Binoculars~\cite{hans2024spotting}, DNA-GPT~\cite{yang2024dnagpt}, and intrinsic dimension estimation~\cite{tulchinskii2023intrinsic}. Despite their diversity, these methods primarily rely on \textit{global aggregation}, which average statistics over the entire sequence and treat token positions as exchangeable, overlooking the critical temporal dynamics of the generation process.


\textbf{Sequential and Temporal Analysis.} While global statistics provide strong baselines, recent work has shifted toward analyzing the dynamic evolution of probability signals. In the frequency domain, Spectrum~\cite{xu2024detecting} and MGT-Prism~\cite{liu2025mgt} reveal that AI-generated text exhibits consistent spectral artifacts in likelihood curves. In the time domain, Lastde~\cite{xu2024training} mines discriminative local subsequences from probability vectors, and DivEye~\cite{basani2025diversity} quantifies the rhythmic diversity of surprisal fluctuations. Theoretical studies have further hypothesized that recursive generation leads to uncertainty contraction over time~\cite{hou2025theoretical}. While these methods may implicitly capture some late-stage signals, for instance, low-frequency components could encode long-range patterns, they do not explicitly analyze the temporal dynamics of the second half, nor do they characterize the specific features we identify. Our work provides the first systematic analysis revealing late-stage volatility decay, where AI-generated text exhibits monotonically decreasing variability as context accumulates.



\section{Conclusion}


We investigated \textit{Late-Stage Volatility Decay}, a temporal signature of autoregressive generation wherein AI-generated text exhibits progressively stabilizing token-level statistics while human writing maintains higher variability throughout. We propose simple yet effective features: Derivative Dispersion and Local Volatility for AI-generated text detection. Despite their simplicity, these temporal features achieve state-of-the-art performance on standard benchmarks across different model families, and demonstrate strong complementarity with global methods, particularly excelling on longer texts where late-stage dynamics fully manifest. Our findings suggest that exploiting the temporal structure of autoregressive generation offers a promising direction for robust detection as language models continue to evolve.

\section*{Limitations}

Our method has two main limitations. First, TSD's effectiveness depends on sufficient sequence length for late-stage volatility decay to manifest. As shown in Appendix~\ref{app:position}, performance degrades when starting positions are set too late (e.g., 90\%), leaving insufficient tokens for reliable estimation. This suggests TSD may be less effective on very short texts where the second half contains too few tokens to capture stable temporal patterns. Second, our fusion strategy for TSD+ employs simple score averaging with Fast-DetectGPT. While effective, this approach does not adapt to varying text characteristics. Future work could explore dynamic or learned fusion mechanisms that adjust weights based on sequence length, domain, or confidence estimates from individual detectors.

\bibliography{main}



\appendix
\label{sec:appendix}
\section{MAGE Results}
\label{app:mage_results}

\subsection{Detailed Results}
\label{app:mage_full}

Table~\ref{tab:mage_full} presents the complete detection performance on the MAGE benchmark across all eight generator families.

\begin{table*}[t]
\centering
\small
\setlength{\tabcolsep}{3pt}
\begin{tabular}{ll|cccccccc|c}
\toprule
\multirow{2}{*}{\textbf{Category}} & \multirow{2}{*}{\textbf{Method}} & \multicolumn{9}{c}{\textbf{MAGE}} \\
\cmidrule(lr){3-11}
 & & OpenAI & GLM & Human-Para & FLAN & LLaMA & BigScience & OPT & EleutherAI & Avg \\
\midrule
\multirow{6}{*}{Global} 
& Likelihood & 89.12 & 91.30 & 60.61 & 30.98 & 92.96 & 37.51 & 31.00 & 21.02 & 56.81 \\
& Entropy & 23.08 & 17.98 & 62.83 & 66.65 & 20.30 & \underline{66.22} & \underline{76.39} & \underline{81.76} & 51.90 \\
& Rank & 73.74 & 83.12 & 53.58 & 34.93 & 81.74 & 45.49 & 52.75 & 46.63 & 58.99 \\
& Log-Rank & 87.05 & 90.28 & 53.81 & 31.56 & 91.69 & 37.38 & 32.39 & 23.10 & 55.91 \\
& LLR & 72.90 & 74.24 & 36.21 & 36.17 & 78.09 & 39.97 & 40.17 & 36.52 & 51.78 \\
& Fast-Detect & 76.40 & 80.41 & 61.92 & \textbf{76.29} & 67.35 & \textbf{72.40} & 64.61 & 58.12 & 69.69 \\
\midrule
\multirow{4}{*}{Temporal} 
& Lastde & 74.47 & 83.87 & 62.96 & 66.35 & 72.01 & 64.63 & 72.35 & 69.07 & 70.71 \\
& FourierGPT & 59.31 & 84.88 & 68.32 & 4.72 & 84.39 & 39.60 & 57.55 & 63.97 & 60.34 \\
& Diveye & 87.89 & 92.61 & 62.10 & 34.65 & 93.80 & 52.00 & 67.60 & 66.11 & 69.60 \\
& UCE & 18.81 & 20.02 & \underline{70.60} & \underline{67.05} & 23.48 & 62.86 & \textbf{82.81} & \textbf{87.73} & 54.17 \\
\midrule
\multirow{4}{*}{Ours} 
& DD & 86.75 & 91.94 & 61.98 & 39.60 & 90.76 & 55.71 & 74.84 & 72.16 & \underline{71.72} \\
& LV & \textbf{91.79} & \textbf{94.41} & 67.21 & 34.61 & \textbf{95.45} & 53.38 & 69.01 & 64.20 & 71.26 \\
& TSD & \underline{90.84} & \underline{94.01} & 65.42 & 35.17 & \underline{94.17} & 53.78 & 71.79 & 67.34 & 71.56 \\
& TSD+ & 89.13 & 93.19 & \textbf{72.03} & 55.92 & 88.34 & 64.88 & 72.74 & 65.36 & \textbf{75.20} \\
\bottomrule
\end{tabular}
\caption{Full detection performance (AUROC \%) on MAGE benchmark across all generator families. Human-Para refers to human-paraphrased AI text. DD: Derivative Dispersion; LV: Local Volatility; TSD: Temporal Stability Detection; TSD+: TSD fused with Fast-DetectGPT. Best results are \textbf{bolded}, second best are \underline{underlined}.}
\label{tab:mage_full}
\end{table*}

The MAGE benchmark reveals complementary strengths across detection approaches, with different methods excelling on specific generator families due to architectural and distributional mismatches with the surrogate model.

Our temporal features achieve the strongest performance on generators that are closer to the Llama-3 surrogate in their decoding behavior. Specifically, LV attains 91.79\% on OpenAI, 94.41\% on GLM, and 95.45\% on LLaMA, outperforming all baselines in these categories. Even the simple DD feature remains competitive (86.75\%, 91.94\%, and 90.76\%), suggesting that derivative-based temporal dispersion captures reliable generation dynamics when the surrogate can effectively characterize token-level uncertainty.
\begin{figure}[t]
\centering
\includegraphics[width=1\linewidth]{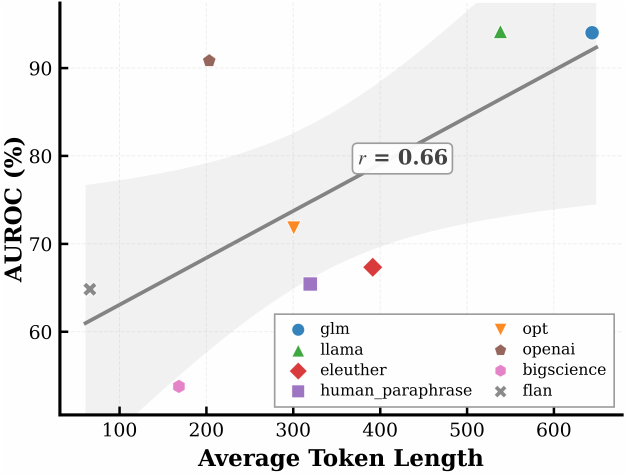}
\caption{Relationship between average token length and temporal detection performance (AUROC) across MAGE generator families. The dashed line shows linear regression fit with Pearson correlation $r = 0.66$}
\label{fig:mage_length}
\end{figure}

The Human-Paraphrase category is particularly challenging because humans revise AI outputs to evade global distributional detectors. Here, TSD+ reaches 72.03\%, outperforming Fast-DetectGPT (61.92\%) by over 10 points. A plausible explanation is that human edits can dilute global cues while still leaving residual late-stage temporal irregularities inherited from the original model output; fusing temporal and global signals mitigates this mismatch.

For generator families that diverge substantially from decoder-only Transformers, global baselines can become more reliable. Fast-DetectGPT achieves the best performance on FLAN-T5 (76.29\%) and BigScience (72.40\%). Importantly, FLAN-T5 samples in MAGE are also typically shorter, which limits the available ``late-stage'' token budget and makes temporal cues less stable. In such cases, Fast-DetectGPT provides a complementary, less length-sensitive signal that compensates for the reduced temporal evidence, lifting TSD from 35.17\% to 55.92\% on FLAN-T5.

Overall, while no single method dominates every family, TSD+ achieves the best average AUROC (75.20\%) by avoiding catastrophic failures and remaining robust across heterogeneous generators. When temporal features are strong (e.g., LLaMA), they substantially improve over global baselines; when temporal evidence is weak (notably for short or architecturally dissimilar generations), the Fast-DetectGPT component acts as a safety net. This complementarity supports our hypothesis that temporal dynamics and global distributional discrepancies capture orthogonal aspects of AI-generated text.

\subsection{Impact of Text Length}
\label{app:mage_length}
\begin{figure}[t]
    \centering
    \includegraphics[width=0.85\linewidth]{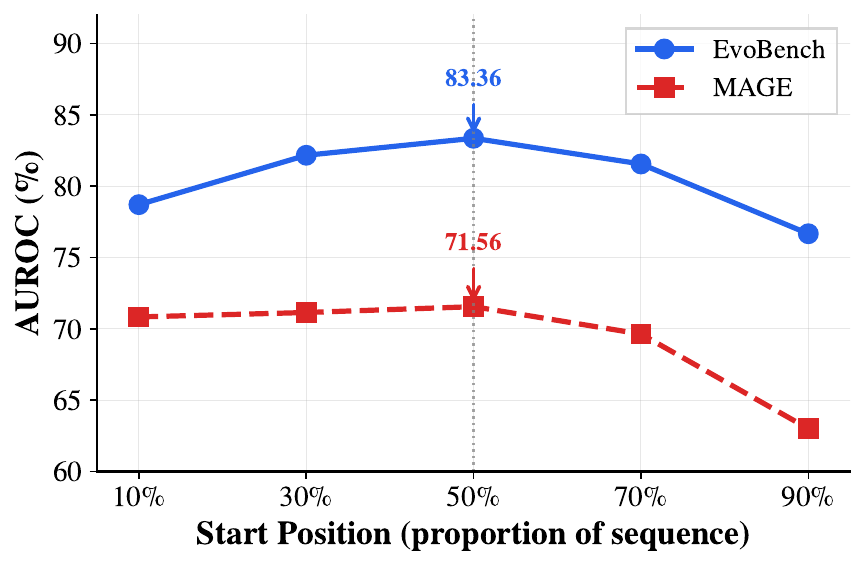}
    \caption{TSD performance versus relative start position. The midpoint (50\%) achieves optimal performance on both benchmarks.}
    \label{fig:position_relative}
\end{figure}

\begin{figure}[t]
    \centering
    \includegraphics[width=0.85\linewidth]{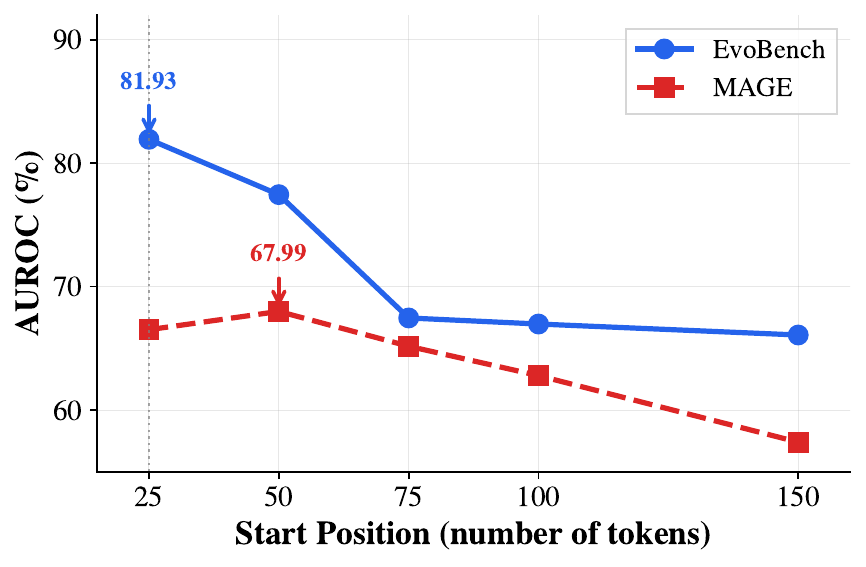}
    \caption{TSD performance versus absolute start position (number of tokens). Performance varies inconsistently across datasets, with EvoBench favoring earlier positions while MAGE shows less sensitivity.}
    \label{fig:position_absolute}
\end{figure}

Figure~\ref{fig:mage_length} shows a clear positive association between average sample length and temporal detection performance on MAGE (Pearson $r=0.66$). This trend is consistent with our design: late-stage temporal cues (e.g., volatility decay / stability patterns) require sufficient token budget to reliably manifest and to be estimated with low variance. Consequently, families with shorter outputs (notably FLAN-T5 in MAGE) tend to exhibit weaker temporal separability, partially explaining why TSD underperforms on FLAN-T5 in Table~\ref{tab:mage_full}. This length sensitivity also motivates our fusion strategy: Fast-DetectGPT contributes a complementary global discrepancy signal that remains effective even when the available temporal evidence is limited, improving robustness across length-diverse generators.

\section{Impact of Start Position}
\label{app:position}

While Section~\ref{sec:method} uses the midpoint (50\%) as the default start position for computing temporal features, we investigate whether alternative positions yield better performance. We conduct experiments using both relative positions (proportion of sequence) and absolute positions (fixed number of tokens).

\paragraph{Relative Position.}
Figure~\ref{fig:position_relative} shows TSD performance as a function of relative start position, ranging from 10\% to 90\% of the sequence. The results reveal an inverted-U pattern on both datasets. Starting too early (10\%--30\%) includes tokens where the late-stage volatility decay has not yet fully manifested, diluting the discriminative signal. Starting too late (70\%--90\%) excludes too many tokens, reducing statistical reliability---particularly evident on MAGE where performance drops sharply from 71.56\% at 50\% to 63.01\% at 90\%. The midpoint achieves optimal performance on both benchmarks: 83.36\% on EvoBench and 71.56\% on MAGE.

\paragraph{Absolute Position.}
We also evaluate using fixed token counts as start positions, shown in Figure~\ref{fig:position_absolute}. While starting at 25 tokens yields high performance on EvoBench (81.93\%), this approach shows inconsistent behavior across datasets---MAGE peaks at 50 tokens (67.99\%) rather than 25 tokens (66.51\%). Moreover, absolute positions fail to adapt to varying sequence lengths: short sequences may have insufficient tokens beyond the start position, while long sequences underutilize available context. In contrast, the relative 50\% position consistently achieves strong performance across both datasets regardless of sequence length distribution.

Based on these findings, we adopt the relative midpoint (50\%) as the default start position, which provides robust performance across diverse datasets without requiring length-specific tuning.

\section{Complete Temporal Dynamics Visualization}
\label{app:mage_vis}

\begin{figure*}[t]
    \centering
    \includegraphics[width=\textwidth]{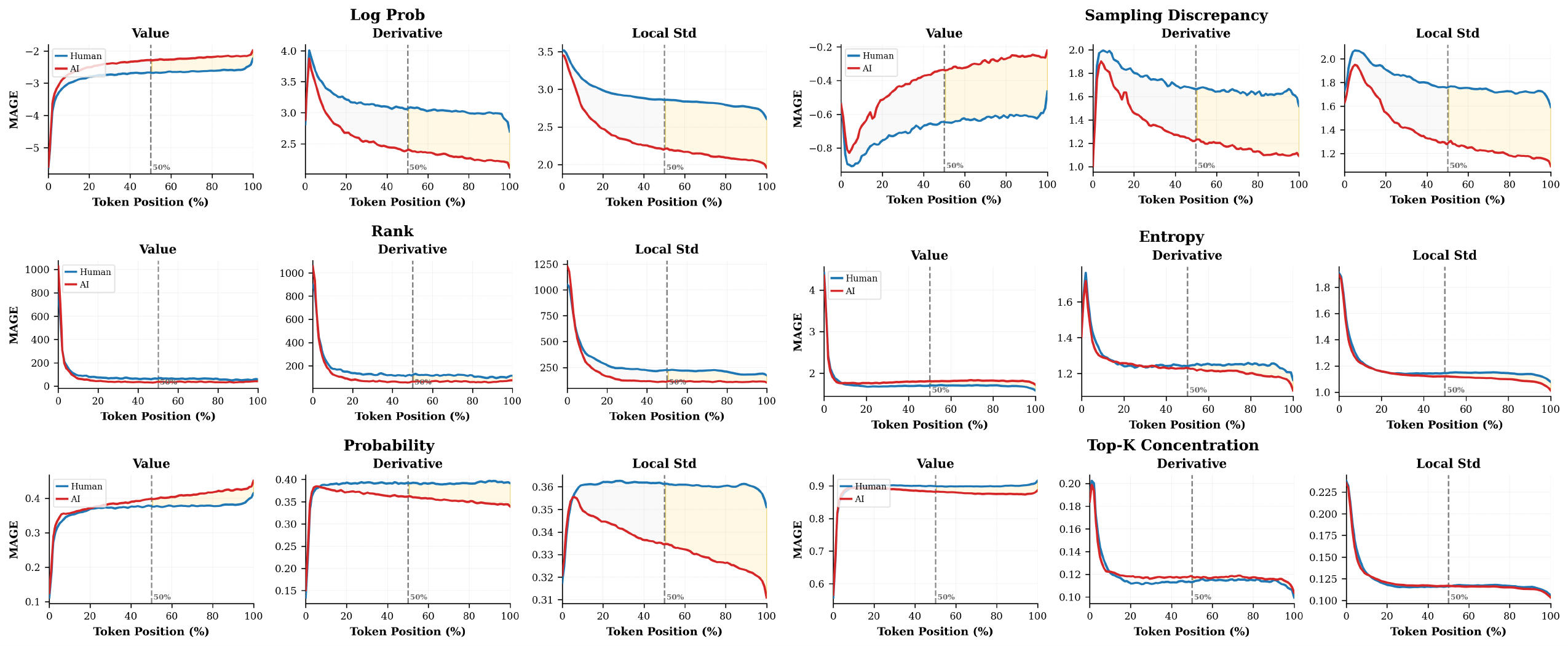}
    \caption{Temporal dynamics of seven token-level metrics on MAGE benchmark. Each metric shows three views: raw value trajectory, derivative (rate of change), and local standard deviation (volatility). The second half of sequences (highlighted in yellow) consistently exhibits larger human-AI gaps in derivative and local standard deviation features, particularly for log probability and sampling discrepancy.}
    \label{fig:mage_temporal}
\end{figure*}

Figure~\ref{fig:mage_temporal} presents the complete temporal dynamics visualization on the MAGE benchmark, complementing the EvoBench results in the main paper. The patterns are consistent across datasets: AI-generated text exhibits steeper decay in both derivative and local standard deviation of log probability, with the gap between human and AI text widening substantially in the second half of sequences.

\section{Surrogate Model Independence}
\label{app:surrogate}

A natural question arises: does the late-stage volatility decay phenomenon depend on the specific choice of surrogate model? To investigate this, we replicate our analysis using two additional surrogate models from different architectural families: GPT-J-6B~\cite{gpt-j} and Falcon-7B~\cite{almazrouei2023falcon}. These models differ substantially from Llama-3-8B-Instruct in training data, architecture details, and model scale, providing a rigorous test of generalization.

\subsection{Temporal Dynamics Across Surrogate Models}

Table~\ref{tab:surrogate_temporal} presents the temporal dynamics of log probability features extracted using each surrogate model on EvoBench. The late-stage volatility decay phenomenon persists across all three surrogate models with remarkably consistent patterns.


\begin{table}[t]
\centering
\small
\setlength{\tabcolsep}{3pt}
\begin{tabular}{ll|cc|cc}
\toprule
\multirow{2}{*}{\textbf{Surrogate}} & \multirow{2}{*}{\textbf{Feature}} & \multicolumn{2}{c|}{\textbf{EvoBench}} & \multicolumn{2}{c}{\textbf{MAGE}} \\
\cmidrule(lr){3-4} \cmidrule(lr){5-6}
 & & Gap\% & Decay & Gap\% & Decay \\
\midrule
\multirow{2}{*}{Llama-3-8B} 
  & Derivative & $-32$ & $2.4$ & $-24$ & $1.9$ \\
  & Local Std & $-31$ & $2.6$ & $-25$ & $1.9$ \\
\midrule
\multirow{2}{*}{GPT-J-6B} 
  & Derivative & $-22$ & $1.8$ & $-20$ & $1.8$ \\
  & Local Std & $-22$ & $2.0$ & $-21$ & $1.9$ \\
\midrule
\multirow{2}{*}{Falcon-7B} 
  & Derivative & $-22$ & $1.9$ & $-19$ & $1.8$ \\
  & Local Std & $-23$ & $2.1$ & $-20$ & $1.8$ \\
\bottomrule
\end{tabular}
\caption{Temporal dynamics of log probability features across different surrogate models. \textit{Gap\%}: relative human-AI difference in the second half. \textit{Decay}: ratio of AI to human decay rate. All surrogate models exhibit consistent late-stage volatility decay patterns.}
\label{tab:surrogate_temporal}
\end{table}

All three surrogate models show: (1) substantial negative gaps in the second half, ranging from $-19\%$ to $-32\%$; (2) AI text decaying $1.8$--$2.6$ times faster than human text. While Llama-3-8B produces the strongest signal (likely due to its closer architectural alignment with recent generators), GPT-J-6B and Falcon-7B still capture meaningful temporal patterns. This consistency suggests that late-stage volatility decay is a fundamental property of autoregressive generation rather than an artifact of specific surrogate-generator pairings.

\subsection{Late-Stage Volatility Decay Details}

Table~\ref{tab:surrogate_late_stage} provides detailed late-stage statistics for each surrogate model, showing the percentage decay from first to second half for both human and AI text.

\begin{table}[t]
\centering
\small
\setlength{\tabcolsep}{2.5pt}
\begin{tabular}{ll|cccc}
\toprule
\textbf{Surrogate} & \textbf{Feature} & \textbf{H-decay} & \textbf{A-decay} & \textbf{Ratio} & \textbf{Gap\%} \\
\midrule
\multicolumn{6}{c}{\textit{EvoBench}} \\
\midrule
\multirow{2}{*}{Llama-3-8B} 
  & Derivative & $-9.6$\% & $-27.1$\% & $2.4$ & $-32$\% \\
  & Local Std & $-8.9$\% & $-26.5$\% & $2.6$ & $-31$\% \\
\midrule
\multirow{2}{*}{GPT-J-6B} 
  & Derivative & $-9.8$\% & $-20.1$\% & $1.8$ & $-22$\% \\
  & Local Std & $-9.1$\% & $-20.3$\% & $2.0$ & $-22$\% \\
\midrule
\multirow{2}{*}{Falcon-7B} 
  & Derivative & $-10.0$\% & $-21.1$\% & $1.9$ & $-22$\% \\
  & Local Std & $-9.3$\% & $-21.9$\% & $2.1$ & $-23$\% \\
\midrule
\multicolumn{6}{c}{\textit{MAGE}} \\
\midrule
\multirow{2}{*}{Llama-3-8B} 
  & Derivative & $-7.5$\% & $-16.7$\% & $1.9$ & $-24$\% \\
  & Local Std & $-7.3$\% & $-16.5$\% & $1.9$ & $-25$\% \\
\midrule
\multirow{2}{*}{GPT-J-6B} 
  & Derivative & $-5.0$\% & $-10.9$\% & $1.8$ & $-20$\% \\
  & Local Std & $-4.9$\% & $-10.9$\% & $1.9$ & $-21$\% \\
\midrule
\multirow{2}{*}{Falcon-7B} 
  & Derivative & $-5.6$\% & $-11.4$\% & $1.8$ & $-19$\% \\
  & Local Std & $-5.3$\% & $-11.0$\% & $1.8$ & $-20$\% \\
\bottomrule
\end{tabular}
\caption{Late-stage volatility decay details across surrogate models. \textit{H-decay}/\textit{A-decay}: percentage decrease from first to second half for human/AI text. The decay ratio remains consistently above $1.8$ regardless of surrogate model choice.}
\label{tab:surrogate_late_stage}
\end{table}

Across all configurations, AI text exhibits substantially faster decay than human text. The decay ratios range from $1.8$ to $2.6$, with human text showing modest decay ($5$--$10\%$) while AI text decays more dramatically ($11$--$27\%$). This asymmetry is precisely what enables our detection method: the autoregressive generation process causes AI text to stabilize toward the end of sequences, while human text maintains more consistent variability throughout.

\subsection{Visualization Across Surrogate Models}

\begin{figure*}[t]
    \centering
    \includegraphics[width=\textwidth]{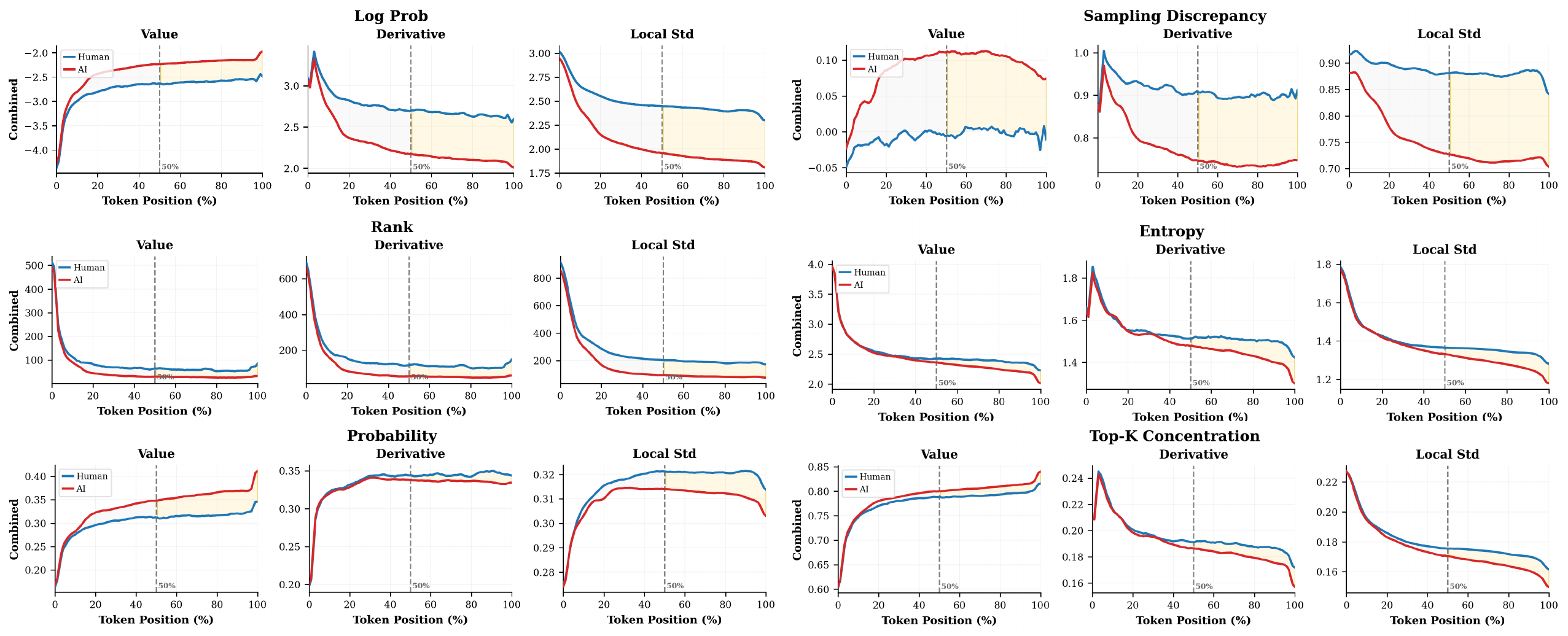}
    \caption{Temporal dynamics on EvoBench using GPT-J-6B as surrogate model. The late-stage volatility decay pattern persists: AI text (red) exhibits steeper decay in derivative and local standard deviation features compared to human text (blue), with pronounced separation in the second half (highlighted in yellow).}
    \label{fig:gptj_temporal}
\end{figure*}

\begin{figure*}[t]
    \centering
    \includegraphics[width=\textwidth]{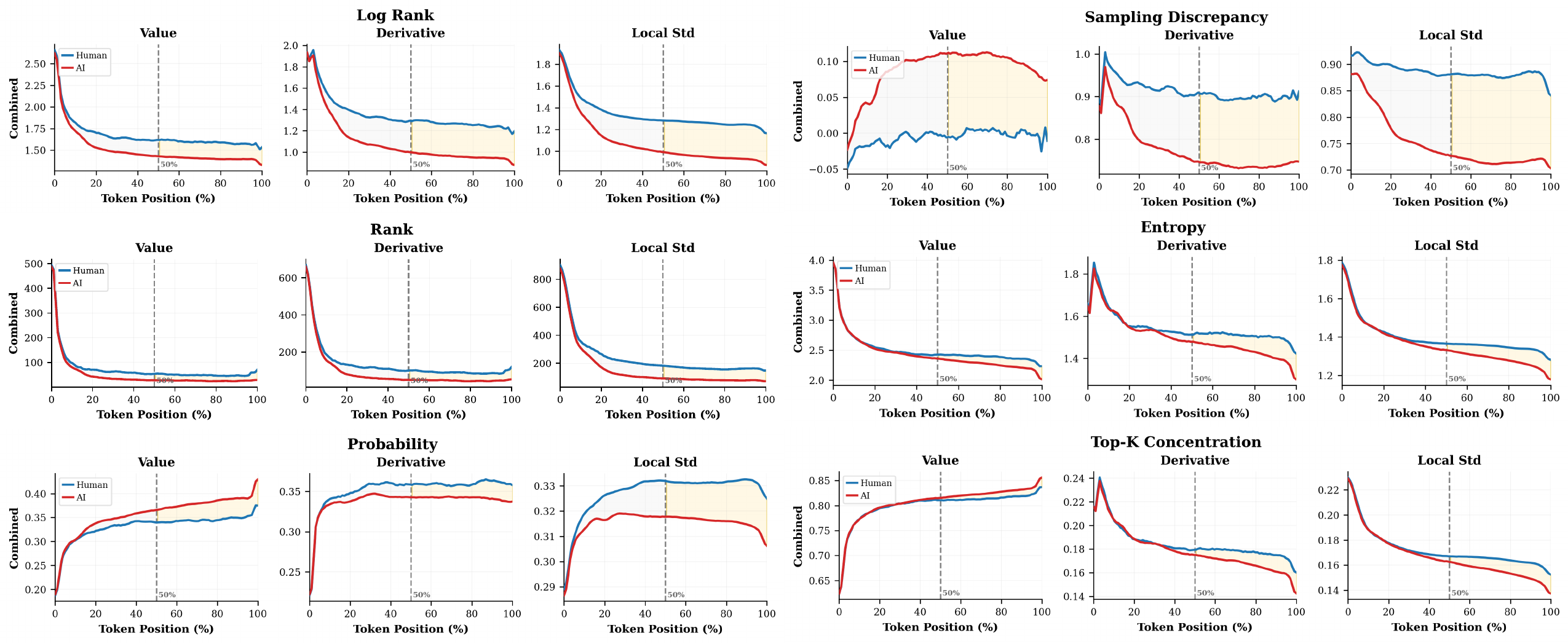}
    \caption{Temporal dynamics on EvoBench using Falcon-7B as surrogate model. Despite different model architecture and training data, the temporal patterns remain consistent with Llama-3-8B and GPT-J-6B results, confirming the surrogate-independent nature of late-stage volatility decay.}
    \label{fig:falcon_temporal}
\end{figure*}

Figures~\ref{fig:gptj_temporal} and~\ref{fig:falcon_temporal} visualize the temporal dynamics using GPT-J-6B and Falcon-7B as surrogate models, respectively. Both figures exhibit the same qualitative patterns observed with Llama-3-8B: log probability shows clear separation in derivative and local standard deviation features, with the gap widening in the second half. Sampling discrepancy similarly maintains strong discriminative power across surrogate models.

The consistency across three diverse surrogate models—spanning different architectures (GPT-style, Llama-style, Falcon-style), scales (6B to 8B parameters), and training procedures—strongly supports our theoretical claim that late-stage volatility decay is an intrinsic property of autoregressive text generation. This surrogate-independence has practical implications: our detection method can leverage whichever surrogate model is most convenient or computationally efficient, without sacrificing the fundamental discriminative signal.

\section{Generalization to Frontier Models}
\label{app:frontier}

To further validate the universality of late-stage volatility decay, we extend our analysis to frontier models released in late 2024 and early 2025, with particular attention to reasoning-enhanced models. We collect text samples from MIRDGE\cite{fu2025detectanyllm}, a recent benchmark containing parallel human-AI pairs across 17 state-of-the-art generators.

\begin{table*}[t]
\centering
\small
\setlength{\tabcolsep}{5pt}
\begin{tabular}{ll|cccc|cccc}
\toprule
\multirow{2}{*}{\textbf{Family}} & \multirow{2}{*}{\textbf{Generator}} & \multicolumn{4}{c|}{\textbf{Derivative}} & \multicolumn{4}{c}{\textbf{Local Std}} \\
\cmidrule(lr){3-6} \cmidrule(lr){7-10}
 & & H-decay & A-decay & Ratio & Gap\% & H-decay & A-decay & Ratio & Gap\% \\
\midrule
\multirow{3}{*}{OpenAI} 
 & GPT-4o & $-10.3$\% & $-26.2$\% & $2.5$ & $-27$ & $-10.4$\% & $-28.3$\% & $2.7$ & $-29$ \\
 & GPT-4o-mini & $-10.5$\% & $-21.8$\% & $2.1$ & $-24$ & $-10.5$\% & $-23.4$\% & $2.2$ & $-26$ \\
 & GPT-o3-mini$^\dagger$ & $-10.4$\% & $-22.5$\% & $2.2$ & $-23$ & $-10.6$\% & $-24.3$\% & $2.3$ & $-25$ \\
\midrule
\multirow{2}{*}{Anthropic} 
 & Claude-3.5-Haiku & $-10.5$\% & $-24.2$\% & $2.3$ & $-24$ & $-10.8$\% & $-26.8$\% & $2.5$ & $-28$ \\
 & Claude-3.7-Sonnet & $-10.5$\% & $-24.8$\% & $2.4$ & $-24$ & $-10.6$\% & $-26.9$\% & $2.5$ & $-27$ \\
\midrule
\multirow{2}{*}{Google} 
 & Gemini-2.0-Flash & $-10.6$\% & $-28.7$\% & $2.6$ & $-24$ & $-10.5$\% & $-30.6$\% & $2.8$ & $-26$ \\
 & Gemini-2.0-Flash-Lite & $-10.6$\% & $-28.8$\% & $2.6$ & $-23$ & $-10.7$\% & $-30.2$\% & $2.8$ & $-26$ \\
\midrule
\multirow{2}{*}{DeepSeek} 
 & DeepSeek-V3 & $-11.0$\% & $-24.4$\% & $1.8$ & $-31$ & $-10.6$\% & $-25.3$\% & $1.9$ & $-32$ \\
 & DeepSeek-R1$^\dagger$ & $-10.0$\% & $-10.9$\% & $1.0$ & $-5$ & $-10.2$\% & $-15.1$\% & $1.4$ & $-11$ \\
\midrule
Meta & Llama-3.1-8B-Instruct & $-11.0$\% & $-29.2$\% & $2.2$ & $-43$ & $-10.7$\% & $-29.2$\% & $2.4$ & $-42$ \\
\midrule
\multirow{2}{*}{Alibaba} 
 & Qwen2.5-7B-Instruct & $-11.0$\% & $-29.6$\% & $2.3$ & $-33$ & $-10.8$\% & $-31.1$\% & $2.4$ & $-35$ \\
 & QwQ-Plus$^\dagger$ & $-10.3$\% & $-14.6$\% & $1.3$ & $-12$ & $-10.7$\% & $-17.9$\% & $1.5$ & $-18$ \\
\midrule
ByteDance & Doubao-1.5-Pro-32K & $-10.0$\% & $-32.2$\% & $3.0$ & $-30$ & $-10.1$\% & $-32.7$\% & $3.0$ & $-30$ \\
\midrule
\multicolumn{2}{l|}{\textit{Average (All)}} & $-10.5$\% & $-24.5$\% & $2.2$ & $-24$ & $-10.6$\% & $-26.3$\% & $2.3$ & $-27$ \\
\multicolumn{2}{l|}{\textit{Average (Reasoning$^\dagger$)}} & $-10.2$\% & $-16.0$\% & $1.5$ & $-13$ & $-10.5$\% & $-19.1$\% & $1.7$ & $-18$ \\
\bottomrule
\end{tabular}
\caption{Late-stage volatility decay across frontier models on MIRDGE. \textit{H-decay}/\textit{A-decay}: percentage decrease from first to second half for human/AI text. \textit{Ratio}: AI-to-human decay ratio. \textit{Gap\%}: relative human-AI difference in the second half. $^\dagger$Reasoning-enhanced models.}
\label{tab:frontier_late_stage}
\end{table*}

Table~\ref{tab:frontier_late_stage} presents the late-stage volatility decay statistics across 13 frontier models. The phenomenon persists universally, with decay ratios ranging from $1.0$ to $3.0$. Critically, we find that late-stage volatility decay extends to reasoning-enhanced models (including GPT-o3-mini, DeepSeek-R1, and QwQ-Plus), which represent a distinct generation paradigm involving extended chain-of-thought processes. While these reasoning models exhibit somewhat attenuated decay ratios at $1.0$--$2.3$ compared to standard models at $1.8$--$3.0$, they nonetheless display the fundamental pattern: AI-generated text stabilizes more than human text toward sequence end. The extended reasoning chains may introduce additional variability that partially masks the decay signal, yet the underlying phenomenon remains detectable.

\begin{figure*}[t]
    \centering
    \includegraphics[width=\textwidth]{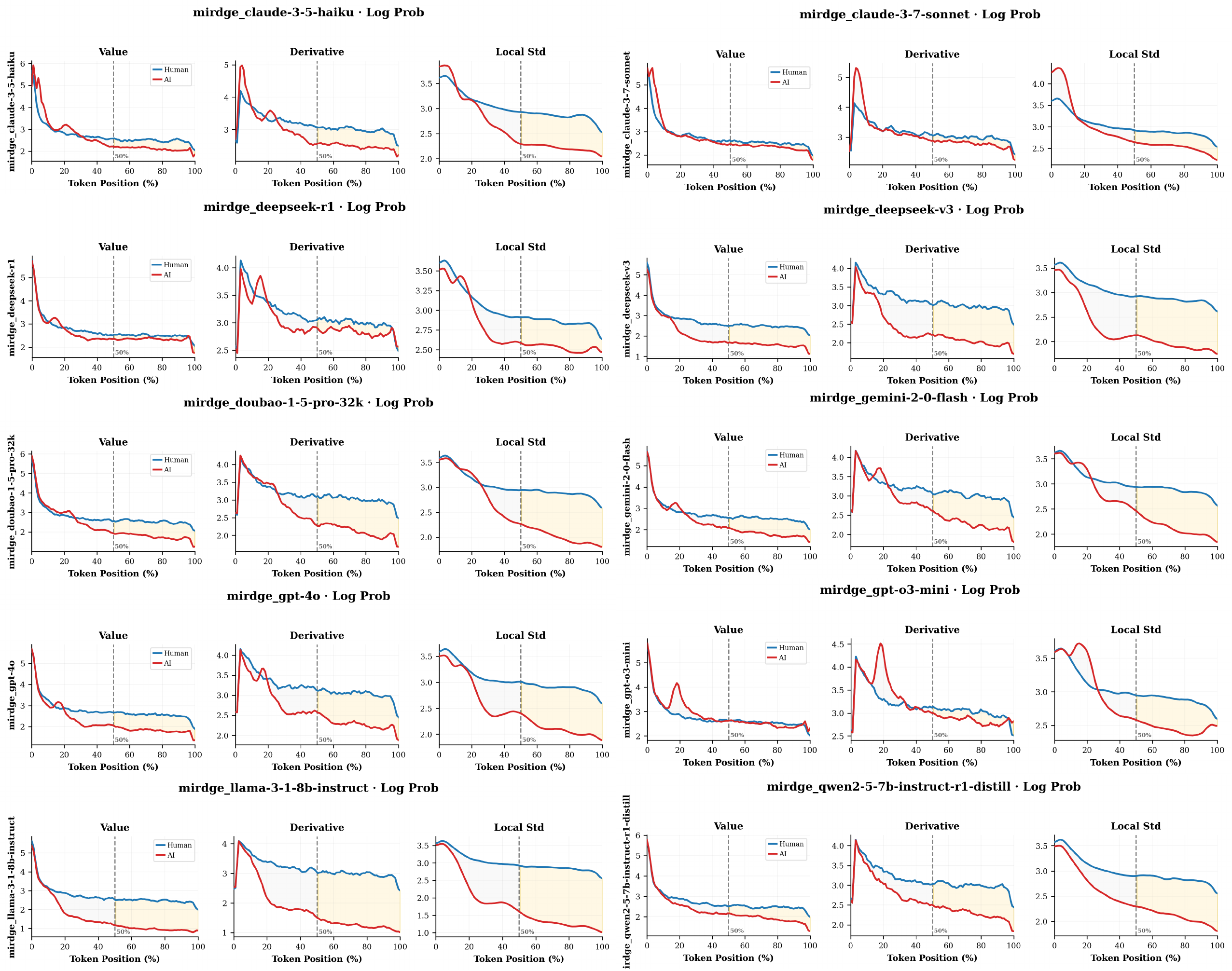}
    \caption{Temporal dynamics of log probability across frontier models from MIRDGE. Each row shows a different generator. The late-stage volatility decay pattern persists universally, including for reasoning models such as DeepSeek-R1 and GPT-o3-mini.}
    \label{fig:mirdge_temporal}
\end{figure*}

Figure~\ref{fig:mirdge_temporal} visualizes the temporal dynamics of a representative subset of models.
Across all model families, AI-generated text consistently exhibits a steeper decay in both the derivative and the local standard deviation, with the separation between AI and human text becoming more pronounced in the latter half of the sequence.
This trend persists regardless of architectural variations or the presence of explicit reasoning mechanisms.

The consistency of late-stage volatility decay across this diverse set of frontier models---spanning different organizations, architectures, scales, and training paradigms---provides strong evidence that this phenomenon is an intrinsic property of autoregressive language generation rather than an artifact of specific model designs.

\section{Additional Statements}

\paragraph{Potential Risks.}
Our method is designed for AI-generated text detection research and should be used responsibly. Like all detection methods, it may produce false positives or negatives and should not be the sole basis for consequential decisions.

\paragraph{Model Size and Computational Budget.}
Our experiments rely on Llama-3-8B-Instruct as the surrogate model for computing token-level statistics. All inference was conducted on one NVIDIA H100 GPU. The choice of surrogate model affects detection performance, and we acknowledge that results may vary with different model sizes or architectures.

\paragraph{Experimental Reproducibility.}
Our approach performs zero-shot detection without requiring training, which ensures stable and deterministic results. All reported results are from single runs, as the absence of stochastic training procedures eliminates the need for multiple runs or variance reporting.

\paragraph{Use of AI Assistants.}
We used AI assistants to aid in manuscript writing and code refinement. These tools helped with language polishing, clarity improvements, and debugging assistance. All core research ideas, experimental design, and analytical conclusions are the authors' own work.

\end{document}